\pdfoutput=1

\documentclass[11pt]{article}

\usepackage{acl}
\usepackage{comment}
\usepackage{tablefootnote}
\usepackage{float}

\usepackage{times}
\usepackage{latexsym}

\usepackage[T1]{fontenc}

\usepackage[utf8]{inputenc}

\usepackage{microtype}

\usepackage{graphicx}

\title{kpfriends at SemEval-2022 Task 2: NEAMER - Named Entity Augmented Multi-word Expression Recognizer
}

\author{Min Sik Oh \\
  Alexa AI\thanks{Research unrelated to work} \\
  \texttt{ohtrent@amazon.com} \\}

\begin{document}
\maketitle
\begin{abstract}

We present NEAMER - Named Entity Augmented Multi-word Expression Recognizer. This system is inspired by non-compositionality characteristics shared between Named Entity and Idiomatic Expressions. We utilize transfer learning and locality features to enhance idiom classification task. This system is our submission for SemEval Task 2: Multilingual Idiomaticity Detection and Sentence Embedding Subtask A OneShot shared task. We achieve SOTA with F1 0.9395 during post-evaluation phase. We also observe improvement in training stability. Lastly, we experiment with non-compositionality knowledge transfer, cross-lingual fine-tuning and locality features, which we also introduce in this paper.
\end{abstract}

\section{Introduction}

Multi-Word Expressions (MWEs) are defined as "idiosyncratic interpretations that cross word boundaries (or spaces)" \cite{10.1007/3-540-45715-1_1}. Recent advances in pre-trained language models such as BERT \cite{devlin2019bert} have enhanced performance of Sentence Classification task, however tasks that specifically identify Multi-Word Expressions (MWE) remain unsolved due to its specific idiomatic properties \cite{garcia-etal-2021-probing,yu-ettinger-2020-assessing}. This SemEval shared task \cite{tayyarmadabushi-etal-2022-semeval} aims to understand Multi-Word Expressions better by novel classification and sentence similarity tasks.

Named Entity Recognition (NER) is a task to identify Named Entities (People, Organizations etc.) in a sentence. Multiple datasets exist that specifically perform this task, including CoNLL-02/03 Shared Tasks for English, German, Spanish and Dutch \cite{tjong-kim-sang-2002-introduction,tjong-kim-sang-de-meulder-2003-introduction}. Multi-Word Expressions and Named Entities are similar in a way that they consist of more than one word but they form a single semantic unit. Thus, Named Entities could be seen as a specific type of Multi-Word Expressions \cite{jackendoff1997architecture,vincze-etal-2011-multiword}. However they are different from idiomatic expressions.

\begin{table}
    \centering
    \begin{tabular}{|p{0.2\columnwidth}|p{0.5\columnwidth}|p{0.15\columnwidth}|}
    \hline
    \textbf{MWE} & \textbf{Target} & \textbf{Label}\\
    \hline
    gold mine & This means that search data is a \textbf{gold mine} for marketing strategy. & 0\newline (Idio-matic)\\
    \hline
    gold mine & The hashtag “Qixia \textbf{gold mine} incident” has been viewed many million of times on the social media site Weibo. & 1\newline(Non-idiom-atic) \\
    \hline
    gold mine & The \textbf{Gold Mine}’s plain frontage \& sparse, white-walled dining room suggest that it’s a quick-fix refuelling stop rather than a place to linger. & 1\newline(Non-idiom-atic)\\
    \hline
    \end{tabular}
    \caption{Dataset samples, table from \cite{tayyar-madabushi-etal-2021-astitchinlanguagemodels-dataset}. Note that 3rd example is a named entity (The Gold Mine referring to a restaurant).}
    \label{tab:oneshot}
\end{table}

We propose \textbf{NEAMER - Named Entity Augmented Multi-word Expression Recognizer} that aim to utilize non-compositionality shared between two streams of NLP research. We explore transfer learning between NER and idiom classification tasks. We also experiment with "locality features" to augment representations of text.

We have participated in Subtask A which is a multilingual classification task to determine if a given sentence has correct idiomatic usage or not. We have focused our efforts on the OneShot setting, where the goal is to classify the target sentence utilizing the ZeroShot dataset consisting of idioms not found in test set and the OneShot dataset consisting of 1 idiom-label pair for all idioms in test set. The dataset has been provided by task organizers \cite{tayyar-madabushi-etal-2021-astitchinlanguagemodels-dataset}.

Contributions of this paper are :

\begin{itemize}
    \item NEAMER system which utilizes transfer learning, NER and other locality features to improve performance and stability of MWE classification task.
    \item Investigation into transfer learning between NER and idiom classification task.
    \item Performance and error analysis to understand capabilities of transfer learning, cross-lingual fine-tuning and locality features.
\end{itemize}

\section{Methodology}
\subsection{Idiom and Named Entity}
\label{sec:idiom_and_named_entity}
Idioms and named entities are similar in the way that when they are comprised of multiple words, collocated words encode extra semantics while individual words lose their semantics partially or completely. This property is referred as non-compositionality \cite{baldwinandkim2010}. "In a nutshell" means "very briefly, giving only the main points" \cite{cb:nutshell} as an idiom; individual words lose their concrete semantics and only the combination specifies intended meaning. Similarly, "Papa John's" refers to "an American pizza restaurant chain" \cite{wiki:papajones} when used as a named entity; in this case, even grammatical functions of individual words are mostly ignored. This similarity is the basis for the transfer learning experiments we performed.

We have discussed similarities, but what about differences? Idioms and named entities refer to completely different usage of MWEs. Idioms are utilized to improve fluency and understandability, or make language more colloquial \cite{baldwinandkim2010}. Named entities are utilized to specify name of persons, organizations and locations \cite{tjong-kim-sang-de-meulder-2003-introduction} and do not have such social purpose. Correspondingly we can expect certain knowledge to be easily transferable between two tasks, while it may take more epochs to obtain best final performance due to fundamental difference between tasks leading to necessity for "unlearning" the previous fine-tuned task. We explore the ideas in the experiments.

\subsection{Transfer Learning and Stability}
\label{sec:stability}

As discussed in Section \ref{sec:idiom_and_named_entity}, idioms and named entities show similar non-compositionality. Thus this is the basis for our transfer-learning experiments, where large language models finetuned on NER task are further trained on idiomatic expression classification task. We investigate following ideas in the experiments:

1. We hypothesize that disparity between task types can bring instability. Large language models are known to be unstable during training \cite{variability2019, instability2020}. Language models are trained using Masked LM pre-training task. The aim of the Masked LM task is to classify every masked word to original word, which results in classification of each tokens to 30,000 possible labels. In contrast, the task at hand is much simpler, with the aim being to classify whole sentence into 2 labels according to usage of relevant MWE. NER task can bridge this task complexity gap since the aim is to classify each tokens to 9 labels.

2. We hypothesize that non-compositionality understanding of the model can be shared between tasks. NER systems need to understand non-compositionality to correctly predict B-XXX tags. It also predicts multiple named entities per sentence. Thus we assert that enough non-compositionality understanding is learnt during the NER fine-tuning process compared to Masked LM task where each token is predicted independently.

We additionally hypothesize that language-specific knowledge could be improved for the model through fine-tuning with similar language data, which we perform experiments on.

\subsection{Locality Features}

We design 5 features that are closely related to MWE usage types. Those are the following:

1. Entity - Whether an MWE contains an NER output span, or an NER output span contains an MWE.

2. Capitalization - Whether any word in the MWE is intentionally capitalized (excluding the first word in a sentence and the case where MWE itself is explicitly capitalized in the dataset).

3. "Be a *" - Whether the MWE starts with a be-verb and the article 'a/an'. Same for Portuguese.

4. "The *" - Whether the MWE starts with "the".

5. Quotation - Whether the MWE is surrounded by quotation marks (" or ').

We name them "locality features" because they expand upon specific position of an MWE by looking at adjacent characters. We encode locality features using a deep neural network to give enough significance to the features during training / inference while enabling them to learn complex relationships between the text. This is further informed by label imbalance (excluding "The *" label, which is balanced) shown in Table \ref{tab:sample_classification}. We perform experiments on whether or not locality features improve the performance on the idiom classification task.

\begin{table}
    \centering
    \begin{tabular}{p{0.25\columnwidth}|p{0.15\columnwidth}|p{0.2\columnwidth}|p{0.2\columnwidth}}
    \hline
    \textbf{Feature} & \textbf{Total} & \textbf{0 (Idiomatic)} & \textbf{1 (Not-idiomatic)} \\
    \hline
    All & 4491 & 2535 & 1956 \\
    \hline
    "The *" & 720 & \textbf{366} & \textbf{354} \\
    Entity & 650 & 94 & \textbf{556} \\
    Capitalized & 634 & 50 & \textbf{584} \\
    Quotation & 165 & \textbf{124} & 41 \\
    "Be a *" & 80 & \textbf{68} & 12 \\
    Parenthesis & 6 & \textbf{5} & 1 \\
    \hline
    \end{tabular}
    \caption{Label statistics in ZeroShot data}
    \label{tab:sample_classification}
\end{table}

\section{Experiment Setup}
\subsection{Model Selection}
\label{sec:model_select}

Experimental results on English ZeroShot (shown in Table \ref{tab:model_selection}) were used to determine pre-trained checkpoints with best performance. We thus selected XLM-Roberta-Large \cite{conneau2020unsupervised} as a starting point for training OneShot models.

The list of checkpoints is: xlm-roberta-base, xlm-roberta-large, xlm-roberta-large-finetuned-conll03-english, xlm-roberta-large-finetuned-conll02-spanish, xlm-roberta-large-finetune-conll03-german, Davlan/xlm-roberta-base-ner-hrl, Davlan/xlm-roberta-large-ner-hrl.

\begin{table}
    \centering
    \begin{tabular}{c|c}
    \hline
    \textbf{Model} & \textbf{ENG F1} \\
    \hline
    mBERT-base (baseline) &  70.7 \\
    xlm-roberta-base & 75.5\\
    xlm-roberta-large & \textbf{79.0} \\
    \hline
    \end{tabular}
    \caption{English ZeroShot F1 on validation data}
    \label{tab:model_selection}
\end{table}

\subsection{Model Architecture}
\label{sec:model_arch}

\begin{figure}
\includegraphics[width=\columnwidth]{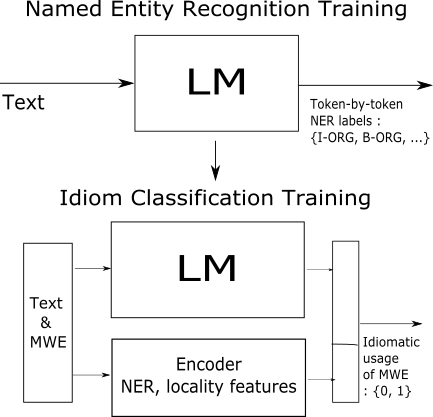}
\caption{NER augmented model, see Section \ref{sec:model_arch} for details.}
\label{fig:ner_model_diagram}
\end{figure}

Our model training scheme and architecture is presented in Figure \ref{fig:ner_model_diagram}. We fine-tune the model on NER task with selected language. For the experiments, we utilize NER fine-tuned checkpoints as described in Section \ref{sec:model_select} instead of actually performing NER fine-tuning. Then, we train the NER fine-tuned model with text and idiom (MWE) data for the idiom classification task along with selected locality features. We use two layers of fully connected network to encode locality features that are concatenated to the text representation. Locality features used are described in Section \ref{sec:locality} and implemented in Python to obtain one-hot vectors which are fed into the fully connected network. The feature encoding and hidden layers of FCN are of size 200. In comparision, LM text encoding is 768 as originally used by XLMRobertaForSequenceClassification class in HuggingFace. The size of encoder feature representation is selected to enhance importance of locality features in comparison to LM representation. We use the classification head provided by the same XLMRobertaForSequenceClassification class.

\subsection{Training Procedure}
\label{sec:train_proc}

We mostly focus on OneShot setting, using both ZeroShot and OneShot data provided. We used a learning rate of $2 \times 10^{-5}$ and a batch size of 16 for training our models. Models were trained for 24 epochs and the best checkpoints on the evaluation data were selected. Random seeds of 0, 1, 3, 5, 42 are used for initial experiments. If any of the seeds exhibit training failures due to instability (F1 < 0.5), we perform additional experiments with random seeds 49, 81, 100, 121. This resulted in at least 5 checkpoints for our experiments. All provided training data was used for training the models. We picked checkpoints that perform best on respective languages (EN / PT) for evaluation and submission. \footnote{Galician test data was inferred by Portuguese model for submission.} We implemented our models in HuggingFace \cite{DBLP:journals/corr/abs-1910-03771} and Pytorch \cite{NEURIPS2019_9015}. We utilize Tesla V100 NVIDIA GPU for training.

\section{Results}

\subsection{Model Stability}
\label{sec:res_stabiltiy}

We present observed training success rate for each of the models in Table \ref{tab:stability}. We define training failure as an observance where F1 of the checkpoint is smaller than 0.5. We observe a very high training failure rate for the XLM-R\textsubscript{large} model (44.4\%). We assert that this is due to discrepancy between the pre-training task of MaskedLM and the idiom classification task (more discussion in Section \ref{sec:stability}.)

\begin{table}[t]
\begin{tabular}{c|c}
\hline
\textbf{Model} & \textbf{Success} \\
\hline
XLM-R & 55.6\% \\
XLM-R-EngNER & 100\% \\
XLM-R-GermanNER & 88.9\% \\
XLM-R-EngNER, Augmented & 100\% \\
\hline
\end{tabular}
\caption{Model training success percentage.}
\label{tab:stability}
\end{table}

\subsection{Best Submissions}

We show our best submissions in Table \ref{tab:comp_res}. Our best official submission during evaluation phase is ensemble of 3 checkpoints per language consisting of XLM-R\textsubscript{large}-EngNER \& SpaNER, with exception of one XLM-R\textsubscript{base}-EngNER checkpoint\footnote{The checkpoints were selected according to best performance on validation set.}. Best post-evaluation submission is ensemble of 5 checkpoints per language consisting of XLM-R\textsubscript{large}-EngNER \& SpaNER, selected via process described in Section \ref{sec:train_proc}. We achieved top 2 during the competition (Section \ref{sec:rank}). We are currently first place in the post-competition leaderboard (4/15/2022).

\begin{table}
    \begin{tabular}{c|c|c|c|c}
    \hline
    \textbf{Phase} & \textbf{ALL} & \textbf{EN} & \textbf{PT} & \textbf{GL} \\
    \hline
    Baseline & 87.7 & 88.1 & 87.0 & 85.4 \\
    Evaluation & 93.5 & \textbf{96.1} & 89.9 & 92.1 \\
    Post-Evaluation\tablefootnote{Experiment performed after end of competition.} & \textbf{94.0} & \textbf{96.1} & \textbf{91.1} & \textbf{92.8} \\
    \hline
    \end{tabular}
    \caption{Best submissions.}
    \label{tab:comp_res}
\end{table}

\subsection{Ensemble Model Performance}

We submit our models based on the ensemble model performance shown in Table \ref{tab:ensemble_perf}. Checkpoints for ensemble were selected via the process described in Section \ref{sec:train_proc}. XLM-R\textsubscript{large} + NER models (xlm-roberta-large-finetuned-conll03-english, xlm-roberta-large-finetuned-conll02-spanish) that represent transfer learning characteristics perform best, with high F1 score across all languages. Interestingly, locality feature augmentation does not seem to enhance the final output compared to the transfer learning only method. This could be due to model checkpoints not having enough variance between them caused by over-reliance on label imbalance. (More discussion in Section \ref{sec:locality})

\begin{table}
    \begin{tabular}{c|c|c|c|c}
    \hline
    \textbf{Model} & \textbf{ALL} & \textbf{EN} & \textbf{PT} & \textbf{GL} \\
    \hline
    XLM-R & 92.7 & 94.5 & 89.5 & 92.3 \\
    XLM-R\textsubscript{NER\textsubscript{HRL, 36}} & 92.5 & \textbf{96.1} & 88.4 & 90.3 \\
    XLM-R\textsubscript{NER\textsubscript{ENG, SPA}} & \textbf{94.0} & \textbf{96.1} & \textbf{91.1} & \textbf{92.8} \\
    \hline
    XLM-R\textsubscript{NER}\textsuperscript{Aug} & 92.8 & 95.6 & 89.4 & 90.8 \\
    \hline
    \end{tabular}
    \caption{Test data F1 performance for ensemble models. All XLM-R models are large variant.}
    \label{tab:ensemble_perf}
\end{table}

\begin{figure}
\centering
\includegraphics[width=1.0\columnwidth]{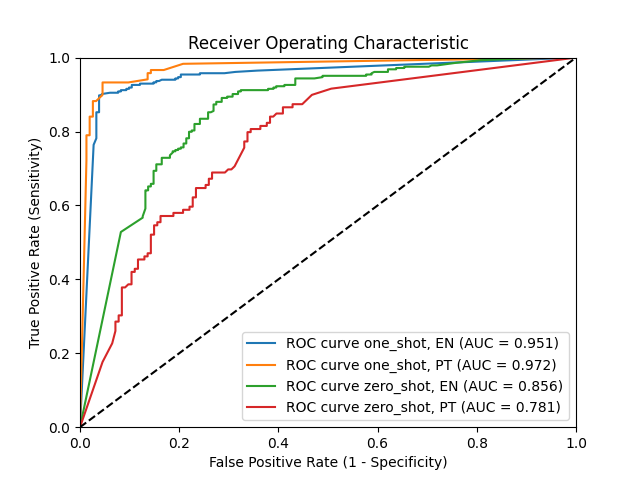}
\caption{ROC curve of XLM-R\textsubscript{NER} on validation data for all tasks. We observe very strong prediction ranking capability for both EN and PT (AUC > 0.950) for OneShot task.}
\label{fig:roc_curve}
\end{figure}

\subsection{Average Model Performance}

The average F1 scores are presented in Table \ref{tab:oneshot_perf}. We observe that additional finetuning on English NER data results in higher performance compared to the baseline XLM-R\textsubscript{large} model. Augmentation of the model using locality features results in a slight performance increase. Results suggest that NER fine-tuning assists in the idiom classification task, while locality features help relatively less. NER fine-tuning is helpful due to the language model adapting to the non-compositionality expressed in both tasks (more discussion in Section \ref{sec:stability}.)

\subsection{Locality Features}
\label{sec:locality}

Effect of locality features seem to be marginal, since average F1 (Table \ref{tab:oneshot_perf}) only slightly improves in comparison with transfer-learning only model. We also observe lower ensemble performance (Table \ref{tab:ensemble_perf}). An enhanced architecture (attention layer in which features explicitly interact with each other) with layer-wise learning rate tuning (to lessen the adverse impact of a cold-start of the feature encoding layers) and dropout (to randomize model training for ensemble enhancement) might be beneficial. We leave it to future work.

We hypothesize that while locality features may be a promising feature to utilize for enhanced architectures, using it by itself may be a relatively too simple indicator. Locality features only require looking at 1\textasciitilde 2 specific tokens\footnote{i.e. Capitalization - first letter of words in MWE, Quotation - ' or " before and after MWE. Parenthesis - ( or ) before and after MWE.}, thus non-compositionality expressed between the tokens themselves is very simple compared to complexity of MWE. An explicit NER feature may also be already encoded in the model via NER fine-tuning step such that no new information is provided during training.

Lastly, we note that we achieve the best ZeroShot setting performance in our experiments with XLM\textsubscript{NER}\textsuperscript{Aug} model which is an ensemble of 3 checkpoints (Table \ref{tab:zeroshot_perf}). Thus, the locality features could be more promising in the ZeroShot setting where there is less information regarding specific MWE usage. We leave a thorough evaluation to future work.

\begin{table}
\begin{tabular}{c|c|c}
\hline
\textbf{Model} & \textbf{Average} & \textbf{Ensemble}\\
\hline
XLM-R\textsubscript{large} & 93.0 & 94.5 \\
XLM-R\textsubscript{large}-Eng & \textbf{94.0} & \textbf{96.1} \\
XLM-R\textsubscript{large}-Eng, Aug & \textbf{94.2} & 95.6 \\
\hline
XLM-R\textsubscript{base}-HRL & 91.1 & - \\
XLM-R\textsubscript{large}-HRL & 92.9 & - \\
XLM-R\textsubscript{large}-HRL, 36 & \textbf{94.2} & \textbf{96.1}\\
\hline
\end{tabular}
\caption{English test data F1}
\label{tab:oneshot_perf}
\end{table}

\begin{table}
    \begin{tabular}{c|c|c|c|c}
    \hline
    \textbf{Model} & \textbf{ALL} & \textbf{EN} & \textbf{PT} & \textbf{GL} \\
    \hline
    XLM-R\textsubscript{NER} & 62.3 & 70.8 & 67.7 & 44.4 \\
    XLM-R\textsubscript{NER}\textsuperscript{Aug} & \textbf{64.9} & \textbf{72.6} & 67.4 & \textbf{49.2} \\
    \hline
    \end{tabular}
    \caption{ZeroShot ensemble test data F1 performance. We note comparatively higher performance for locality feature augmented model on English and Galician data.}
    \label{tab:zeroshot_perf}
\end{table}

\subsection{Crosslingual NER Transfer Learning}

XLM-R\textsubscript{large}-HRL is an XLM-R\textsubscript{large} model trained on NER tasks for 10 languages (Arabic, German, English, Spanish, French, Italian, Latvian, Dutch, Portuguese and Chinese). Rationale for fine-tuning this model is to observe the following :

1. Impact of fine-tuning on a model from a pre-trained model trained on NER data from multiple languages. This model has been trained on all CONLL02 / 03 datasets for English, Spanish, Dutch and German, as well as 8 language specific datasets.

2. Impact of fine-tuning on a model which has been pre-trained with capability to perform Portuguese NER task. This model has been trained on Paramopama and Second Harem \cite{harem-2010} Portuguese NER datasets.

\begin{table}
    \begin{tabular}{c|c|c|c}
    \hline
    \textbf{Model} & \textbf{EN} & \textbf{PT} & \textbf{GL} \\
    \hline
    XLM-R\textsubscript{large}-Eng, Spa & 94.0 & 87.5 & 88.5 \\
    XLM-R\textsubscript{large}-German & 93.6 & 87.2 & 84.2 \\
    \hline
    XLM-R\textsubscript{base}-HRL & 91.1 & 83.6 & 83.2 \\
    XLM-R\textsubscript{large}-HRL & 92.9 & 84.0 & 83.7 \\
    XLM-R\textsubscript{large}-HRL, 36 & \textbf{94.2} & 85.9 & 87.2 \\
    \hline
    \end{tabular}
    \caption{Test data average F1 performance for HRL model variants and English, Spanish and German NER fine-tuned model.}
    \label{tab:hrl_perf}
\end{table}

We show the results in Table \ref{tab:hrl_perf}. We observe that while XLM-R\textsubscript{large}-HRL performs worse on EN F1 than the similarly fine-tuned XLM-R\textsubscript{large}-English and German, training for 36 epochs (50\% epoch increase) yields comparable performance. This aligns with our hypothesis that task-to-task training requires "unlearning" partial aspects of the previous task and thus may take longer to train (more discussion in Section \ref{sec:idiom_and_named_entity}). XLM-R\textsubscript{large}-English was only trained on CoNLL03 English NER task, while HRL models were trained on NER datasets corresponding to 10 languages - this may result in a higher amount of NER task and language specific knowledge that needs to be removed for the model to train properly.

Similarly, we observe worse performance on Portuguese and Galician results for HRL models compared to Spanish fine-tuned model. Portuguese and Galician seem to require more training epochs than English to achieve comparable performance. This may be due to the difference in dataset size per language in both the ZeroShot and OneShot training data for idiom classification task (English:Portuguese = 2.9:1). We leave training the models on more Portuguese idiom classification datasets and longer epochs to future work.

We also experiment with a model fine-tuned on CoNLL 03 German NER task. We note slightly worse performance for German fine-tuned model compared to models fine-tuned on highly similar languages (English and Spanish NER fine-tuned models). This result seems to suggest that fine-tuning the model on same language for both NER task and Idiom Classification task achieves best performance. More experiments with many languages from other parts of the world could be performed.

\section{Error Analysis}

\subsection{Categorical Performance}
\label{sec:loc_perf}

We show the F1 metrics for the validation data per each feature in Table \ref{tab:f1_metric_val}. We find that the F1 score of "The *" locality feature has increased by 5.8 points after transfer learning is introduced. This locality feature does not directly correspond to NER, and is the only sample-balanced locality feature as shown in Table \ref{tab:sample_classification}. Thus, we argue that this is further proof of NER transfer learning teaching general non-compositionality to LM that is transferred to MWE classification task.

We also find that Capitalized and Entity F1 scores have stayed the same after the introduction of NER transfer learning, and it has actually decreased by 2\textasciitilde 3 points after locality feature augmentation. We also observe a recall decrease of 0.214 (0.357 -> 0.143) as shown in Table \ref{tab:entity_conf}. As discussed in Section \ref{sec:locality}, this is due to over-reliance on training data label imbalance.

\begin{table}
\centering
    \begin{tabular}{p{0.35\columnwidth}|p{0.15\columnwidth}|p{0.15\columnwidth}|p{0.15\columnwidth}}
    \hline
    \textbf{Feature} & \textbf{LM} & \textbf{NER} & \textbf{Aug} \\
    \hline
    Capitalized (137) & 94.2 & 94.2 & 91.2 \\
    Entity (131) & 93.1 & 93.1 & 90.8 \\
    \hline
    "The *" (52) & 86.5 & \textbf{92.3} & \textbf{92.3} \\
    \hline
    "Be a *" (13) & 100.0 & 100.0 & 100.0 \\
    Quoted (12) & 90.5 & 90.5 & 90.5\\
    \hline
    \end{tabular}
    \caption{Micro F1 Metrics (validation data) for each locality feature tagged samples corresponding to XLM-R, XLM-R\textsubscript{NER} and XLM-R\textsubscript{NER}\textsuperscript{Aug}. We observe that transfer learning has improved the performance for "The *" feature. More discussion in Section \ref{sec:loc_perf}.}
    \label{tab:f1_metric_val}
\end{table}

\begin{table}
    \centering
    \begin{tabular}{c|c|c|}
     & Pred 0 & Pred 1 \\
    \hline
    Label 0 (Idiomatic) & 5 & 9 \\
    \hline
    Label 1 (Non-idiomatic) & 0 & 117\\
    \hline
    \end{tabular}
    \newline
    \vspace*{1 em}
    \newline
    \centering
    \begin{tabular}{c|c|c|}
     & Pred 0 & Pred 1 \\
    \hline
    Label 0 (Idiomatic) & 2 & 12 \\
    \hline
    Label 1 (Non-idiomatic) & 0 & 117\\
    \hline
    \end{tabular}
    \caption{Confusion matrix for Entity in non-augmented models(XLM-R, XLM-R\textsubscript{NER}) vs augmented model (XLM-R\textsubscript{NER}\textsuperscript{Aug}).}
    \label{tab:entity_conf}
\end{table}

\subsection{Sample Analysis}

We list the prediction improvements between base XLM-R\textsubscript{large} model and NER transfer-learning based models in Appendix \ref{sec:appendix_correct}. Interestingly, we observe that 6 out of 9 sample prediction improvements for English model are also observed with HRL, German\footnote{German model is not trained on CoNLL03 English data, making the result more interesting.} models. This strongly suggests that shared characteristics are present between NER transfer-learning based models. We also observe that the model output changes are not associated with named entities, strengthening our hypothesis of general non-compositionality knowledge transfer between tasks.

\section{Conclusion}

We present NEAMER - Named Entity Augmented Multi-word Expression Recognizer. This system explores how we can utilize non-compositionality shared between Named Entity and Idiomatic Expressions. We find that the NER transfer learning variant achieves the best MWE classification OneShot performance. We also observe high training stability. We investigate non-compositionality knowledge transfer between tasks and obtain promising results across experiments.

\section{Rank Information}
\label{sec:rank}

During the official evaluation phase, we were top 2 in Subtask A (One-Shot) leaderboard with F1 score of 0.9346 (Table \ref{tab:comp_res}). We trained 50 checkpoints and measured F1 on English and Portuguese separately. Checkpoints were generated via process described in \ref{sec:train_proc}. Best English performing checkpoints inferred on English test submission data, while best Portuguese performing checkpoints inferred on Galician as well as Portuguese test submission data. Finally, we ensembled best performing models on each language using different strategies (including top 3, top 5, top 10) to optimize generalization performance.

\section{Acknowledgements}

We'd like to sincerely thank SemEval Task 2 Organizers, especially Harish Tayyar Madabushi, for organizing the Shared Task and providing valuable correspondence during review. We also thank Young Sun Yoon for sharing his knowledge of Ibero-Romance languages.

\bibliography{anthology,custom}
\bibliographystyle{acl_natbib}

\appendix

\section{Prediction Improvements}
\label{sec:appendix_correct}

We list the classification improvements\footnote{Wrong prediction in XLM-R\textsubscript{large} model, but correct prediction in NER transfer learning models.} in validation dataset observed across NER transfer learning models in comparison to base XLM-R\textsubscript{large} model. The NER transfer learning models we compare are English, German, and HRL (10 languages). We find 6 samples that prediction have improved consistently across all 3 models, which is 66.7\% of prediction improvements in English model.

\begin{table}[H]
\centering
\begin{tabular}{p{0.15\textwidth}|p{0.65\textwidth}|p{0.1\textwidth}}
\hline
\textbf{MWE} & \textbf{Sentence} & \textbf{Feature}\\
\hline
high life & "This is the story of “Memo Fantasma” or “Will the Ghost,” who started life in the Medellín Cartel, funded the bloody rise of a paramilitary army, and today lives the high life in Madrid." & "The *"\\
\hline
home run & He is the only player to hit at least 30 home runs in 15 seasons and is one of only four players to produce at least 17 seasons with 150 or more hits. & - \\
\hline
health check & Big Tech Show · Why your DNA may be your next health check & - \\
\hline
pillow slip & By morning most of it is on the pillow slip, and soap and water will clean up the rest." & "The *" \\
\hline
pillow slip & "Her pillow slip by now was very much askew; one ear pointed northward, the other southeast, and she could only see out of one eye." & - \\
\hline
dry land & And God called the dry land Earth; and the gathering together of the waters called he Seas: and God saw that it was good. & "The *" \\
\hline
\end{tabular}
\caption{Improved samples due to NER fine-tuning.}
\label{fig:validation-data-improv}
\end{table}

\end{document}